\documentclass[conference]{IEEEtran}
\IEEEoverridecommandlockouts
\usepackage{cite}
\usepackage{amsmath,amssymb,amsfonts}
\usepackage{algorithmic}
\usepackage{graphicx}
\usepackage{textcomp}
\usepackage{xcolor}
\usepackage{tikz}
\usepackage{url}
\usepackage{multirow} 
\usetikzlibrary{mindmap}

\def\BibTeX{{\rm B\kern-.05em{\sc i\kern-.025em b}\kern-.08em
    T\kern-.1667em\lower.7ex\hbox{E}\kern-.125emX}}
\begin{document}

\title{Utilizing Large Language Models for Information Extraction from Real Estate Transactions
}

\makeatletter
\newcommand{\linebreakand}{%
  \end{@IEEEauthorhalign}
  \hfill\mbox{}\par
  \mbox{}\hfill\begin{@IEEEauthorhalign}
}
\makeatother

\author{

\IEEEauthorblockN{1\textsuperscript{st} Yu Zhao \IEEEauthorrefmark{1}\thanks{* Corresponding author.}}
\IEEEauthorblockA{\textit{Rotman School of Management} \\
\textit{University of Toronto}\\
Toronto, Canada \\
yzqr.zhao@mail.utoronto.ca}

\and

\IEEEauthorblockN{2\textsuperscript{nd} Haoxiang Gao} 
\IEEEauthorblockA{\textit{ECE Department} \\
\textit{Carnegie Mellon University}\\
Mountain View, United States \\
haoxiang@alumni.cmu.edu}

\and
\IEEEauthorblockN{3\textsuperscript{rd} Jinghan Cao}
\IEEEauthorblockA{\textit{Department of Computer Science} \\
\textit{San Francisco State University}\\
San Francisco, United States \\
jcao3@alumni.sfsu.edu}
\linebreakand

\IEEEauthorblockN{4\textsuperscript{th} Shiqi Yang}
\IEEEauthorblockA{\textit{New York University}\\
New York, NY United States \\
939137037@qq.com}
}

\maketitle

\begin{abstract}
Real estate sales contracts contain crucial information for property transactions, but manual data extraction can be time-consuming and error-prone. This paper explores the application of large language models, specifically transformer-based architectures, for automated information extraction from real estate contracts. We discuss challenges, techniques, and future directions in leveraging these models to improve efficiency and accuracy in real estate contract analysis. We generated synthetic contracts using the real-world transaction dataset, thereby fine-tuning the large-language model and achieving significant metrics improvements and qualitative improvements in information retrieval and reasoning tasks.
\end{abstract}


\section{Introduction}
Real estate transactions involve complex legal documents, such as sales contracts, that outline the terms and conditions agreed upon by buyers and sellers. 
Extracting key information from these contracts is essential for various purposes, including due diligence, risk assessment, and compliance. However, manual review and extraction of data from these documents can be labor intensive and prone to errors.

Real estate transactions are uniquely complex due to several distinctive factors. One key aspect is the presence of contingencies, which are conditions that must be met for the transaction to proceed, such as financing or inspection contingencies. These contingencies add layers of negotiation and uncertainty. In addition, real estate transactions often involve an executory period that spans weeks or months, allowing time for inspections and repairs before final closing. Property ownership is transferred through a deed, a legal document that conveys ownership rights, and must be carefully drafted and executed. Furthermore, transactions entail various liabilities, such as environmental issues or property defects, requiring disclosure and mitigation to minimize risk. These factors highlight the specialized expertise needed to navigate real estate transactions successfully.

In this paper, we investigate the use of large language models for extracting structured data from real estate sales contracts. Traditionally, LSTM recurrent networks\cite{schmidhuber1997long} and Transformers\cite{vaswani2017attention} have been widely used to analyze sequential data in a variety of domains. We explore techniques to preprocess documents, fine-tune models, and information extraction techniques. 

\section{Related Work}
Recent advances in natural language processing (NLP) \cite{chang2024survey}, particularly with language models like BERT (Bidirectional Encoder Representations from Transformers) \cite{devlin2018bert} and GPT (Generative Pre-trained Transformer) \cite{2005.14165}, offer promising solutions for automating the extraction of information from text. These models, which have been proven to excel in a number of applications requiring sophisticated context understanding and reasoning\cite{gao2024survey}, excel in understanding and generating human-like text, making them suitable for complex document analysis tasks.

Traditional statistical and machine learning approaches such as Conditional Random Fields (CRF), Support Vector Machines (SVM) and Hidden Markov Model (HMM) have been explored to extract named entities from legal contracts \cite{survey_ner_classification, betts2016dawn, surden2021machine, cui2023chatlaw, zamani2017using}. \cite{location_ml_contracts} applied a sequence of information retrieval and traditional machine learning methods to determine the governing law of a contract. 

While there has been ample research on large language models, the applications of machine learning models to law are scarce. To the best knowledge of the author, the applications of large language models in the domain of real estate transactions have not been well enough studied.

\section{Motivations}
The motivation for employing Legal Language Models (LLMs) in reading and understanding real estate transaction contracts is multifaceted and impactful. One key benefit is the optimization of attorney time. LLMs can swiftly analyze lengthy contracts, identify critical clauses, and flag potential issues, enabling attorneys to focus their efforts on higher-level legal analysis and strategic decision-making. This streamlined approach not only enhances productivity but also ensures that legal professionals can devote more time to addressing complex aspects of the transaction, ultimately providing greater value to their clients.

When combined with inspection reports, appraisal reports, and past transaction history from recorded conveyance records, Large Language Models (LLMs) have the potential to consolidate a wealth of information and streamline the transaction reporting process. This integration allows LLMs to distill key insights and generate concise reports summarizing critical details such as property condition, valuation, ownership history, and legal implications. The ability of LLMs to synthesize disparate information into a cohesive and easily digestible format not only saves time but also enhances decision-making by providing stakeholders with a clear overview of the transaction's key aspects and potential implications. Ultimately, leveraging LLMs alongside other pertinent reports and records enables a more efficient and informed approach to real estate transactions.

\section{Methodology}

To extract information from real estate contracts using large language models, we adopt the following methodology:

\subsection{Data and Feature Preprocessing}

Data processing for real estate transaction contracts within the context of large language models involves several intricate steps leveraging mathematical operations and matrix representations. Initially, raw text representing real estate contracts is tokenized into a sequence of tokens \( x_1, x_2, \dots, x_n \), where each token corresponds to a word or subword unit within the document. These tokens are then mapped to unique indices using a vocabulary mapping function \( \text{vocab}(x_i) = i \), forming a tokenized input sequence \( \mathbf{x} = (x_1, x_2, \dots, x_n) \). Subsequently, tokens are embedded into dense vector representations using an embedding matrix \( \mathbf{E} \), where each token \( x_i \) is represented as a vector \( \mathbf{e}_i = \mathbf{E}[i, :] \). The embedded sequence \( \mathbf{X} = (\mathbf{e}_1, \mathbf{e}_2, \dots, \mathbf{e}_n) \) captures semantic and contextual information crucial for understanding real estate contract text.

To convey sequential information in the input, Rotary Position Embedding (ROPE) \cite{su2024roformer} can be incorporated. ROPE is a relative positional encoding technique that applies a rotation operation to the token embeddings, enabling the model to capture positional relationships directly in the attention mechanism. For a given token embedding \( \mathbf{x} \in \mathbb{R}^d \) at position \( pos \), ROPE applies a rotation matrix \( \mathbf{R}(pos) \), transforming the embedding as follows:
\[
\text{ROPE}(\mathbf{x}, pos) = \mathbf{R}(pos) \cdot \mathbf{x},
\]
where \( \mathbf{R}(pos) \) is a block-diagonal matrix that rotates embedding dimensions in pairs. This approach integrates relative positional information into the model without introducing explicit positional vectors into the embedding layer.

Real estate transactions often exhibit less sequential interdependence, as each paragraph in a contract or report typically corresponds to a distinct clause or specific point. ROPE is well-suited for such data, as it efficiently encodes positional relationships while preserving flexibility for processing relatively independent sections of text. By encoding positional relationships implicitly, ROPE enhances the model's ability to understand both local and global contexts in real estate contracts.

Furthermore, preprocessing steps such as padding, truncation, or batch formation can ensure uniform input dimensions and facilitate efficient training and inference. These mathematical operations and matrix manipulations are fundamental in transforming raw real estate transaction text into structured numerical inputs suitable for processing by large language models, enabling tasks such as contract analysis, information extraction, and legal document understanding.

\subsection{Fine-tuning Large Language Models}

Fine-tuning large language models (LLMs) for real estate contract text data involves various approaches that leverage transfer learning, task-specific fine-tuning, and multi-task learning to enhance model performance and enable domain-specific learning. Transfer learning involves using pre-trained LLMs, like BERT\cite{devlin2018bert} or GPT\cite{achiam2023gpt}, trained on massive datasets for general language understanding. These models are then fine-tuned on real estate contract text data to adapt their learned features and knowledge to the nuances of real estate transactions. The fine-tuning process updates model parameters to better align with the target domain, enhancing the LLM's ability to understand real estate-specific language and concepts.

Task-specific fine-tuning is another approach where pre-trained LLMs are fine-tuned on a dataset tailored specifically for real estate contracts. This method focuses on optimizing the LLM's performance for a particular task, such as contract summarization or clause classification. By training the model on task-specific data, it learns to extract relevant information and make domain-specific predictions, effectively enhancing its ability to process real estate contract text.

Multi-task learning is a technique that involves training an LLM on multiple related tasks simultaneously \cite{mahabadi2021parameter}, including real estate-specific tasks like contract interpretation and clause extraction. This approach has been used in numerous applications in specific domains \cite{chakrabarty2019imho} because it encourages the model to learn shared representations across tasks, enabling it to generalize better and improve performance on each individual task\cite{howard2018universal} \cite{wallingford2022task}. Multi-task learning helps the LLM leverage common patterns and features across different real estate contract-related tasks, leading to enhanced domain-specific learning and more robust performance.

\subsection{Information Extraction}

Another approach involves utilizing sequence labeling models, such as conditional random fields (CRFs), to capture structured information within real estate contracts. CRFs can model dependencies between tokens in text and assign labels to sequences corresponding to different contract elements like contingency clauses or price terms. By training LLMs with annotated real estate contract data and CRFs, the model can learn to identify and extract key information effectively.

Furthermore, LLMs can leverage semantic parsing techniques to understand the semantics of real estate contract text and extract specific attributes like property details, contract conditions, and financial terms. Semantic parsing involves mapping natural language expressions to structured representations, enabling LLMs to interpret complex contract language and extract structured information like property attributes and contractual obligations.

Mathematically, training large language models (LLMs) involves learning to predict the next token in a sequence of real estate contract text \( x \). The dataset \( \mathcal{D} \) consists of labeled real estate contract data, and the goal is to predict key information categories \( \mathcal{C} \) token by token. Let \( f(x; \theta) \) represent the LLM parameterized by \( \theta \), and let the predicted probability distribution over the vocabulary for the next token \( x_{t+1} \) be \( \hat{y}_{t+1} = f(x_{t}; \theta) \).

The training objective is to minimize the negative log-likelihood of the correct token \( x_{t+1} \) under the predicted probability distribution:
\[
\mathcal{L}(\theta) = - \sum_{t} \log P(x_{t+1} \mid x_{\leq t}; \theta),
\]
where \( P(x_{t+1} \mid x_{\leq t}; \theta) \) is obtained via a softmax over the model's logits for the next token.

During generation, the model operates in an autoregressive pattern, where each predicted token \( x_{t+1} \) is conditioned on all previous tokens \( x_{\leq t} \). This process continues until an end-of-sequence token is generated, enabling the extraction of structured information from real estate contracts in a sequential manner.

\section{Query}
Once an LLM model is fine-tuned, it is capable of answering a wide range of questions related to real estate transactions. For example, queries like "Describe the area of the property" can be effectively answered by the model. Similarly, questions about specific contract terms, such as "What are the contingencies in the contract?" can be addressed based on the model's training on legal language and contract structures. By fine-tuning the LLM with relevant real estate data and legal documents, the model gains the ability to interpret and respond to diverse inquiries related to property details, transaction terms, and legal provisions, providing valuable insights and information to users involved in real estate transactions. 

In fact, real estate professionals can create a customized set of questions that encompass key due diligence questions. These questions can be fed into the model for each transaction to automatically extract information for an arbitrary transaction.

\section{Dataset Generation}
 There are a number of challenges that arise when trying to analyze real estate transactions, one of the biggest being the scarcity of data due to privacy concerns. Public researchers have no access to a large volume of real estate transaction contracts between private buyers and sellers, and it's also difficult to acquire the dataset covering all the diversity and complexity of real estate contracts, e.g., differences between different states and property types. The lack of access to a comprehensive dataset of real estate transaction contracts between private buyers and sellers is a significant challenge for public researchers. This limitation hinders their ability to conduct in-depth analysis and derive meaningful insights into the real estate market. Several factors contribute to this data accessibility issue.

Firstly, real estate transactions are often considered private and confidential. As a result, the parties involved are hesitant to share their contracts with third parties, including researchers. This reluctance stems from concerns about privacy, potential legal implications, and the sensitive nature of the information disclosed in the contracts.

Secondly, the real estate market is highly fragmented and diverse. There are significant variations in laws, regulations, and practices across different states and localities. Additionally, the types of properties involved in transactions can range from residential homes to commercial buildings, each with its own unique set of contractual considerations. This diversity makes it challenging to collect a comprehensive dataset covering the entire real estate contract spectrum.

The absence of such a dataset has far-reaching implications for public researchers. It limits their ability to accurately assess market trends, analyze the impact of government policies, and evaluate the effectiveness of various real estate-related interventions. Without access to a comprehensive dataset, researchers are forced to rely on limited or incomplete information, which can lead to biased or inaccurate conclusions.
 
Real-world contracts are in unstructured formats, such as scanned images and PDF formats, making it difficult to convert the raw contracts into structured text data that can be consumed by machine learning models. This lack of data makes it difficult to identify trends and patterns and can lead to inaccurate or biased results.

On top of that, several other challenges arise when trying to analyze real estate transactions. These include:
\begin{itemize}
    \item The complexity of real estate contracts: Real estate contracts are often long and complex and can contain a variety of clauses and conditions that can be difficult to interpret.
    \item The lack of standardization in real estate contracts: There is no standard format for real estate contracts, and the terms and conditions can vary widely from one contract to another. This can make it difficult to compare different contracts and identify trends.
    \item The need for expert knowledge: Analyzing real estate transactions requires a deep understanding of the real estate market and its legal framework. This can make it difficult for non-experts to conduct meaningful analyses.
\end{itemize}

Synthetic data has emerged as a valuable tool for training Large Language Models (LLMs) due to its potential to overcome limitations associated with real-world data, such as privacy concerns, biases, and limited availability of specific examples. Several methods have been explored to leverage synthetic data for LLM training\cite{liu2024best}:

\begin{itemize}
    \item Data Augmentation: Existing datasets are enhanced by generating additional variations of existing samples. This can involve techniques like paraphrasing, text transformations, or generating new examples based on specific patterns.
    \item Rule-Based Generation: Synthetic data is generated based on predefined rules or templates. This approach is useful when specific linguistic patterns or structures must be reinforced.
    \item Model-Based Generation: LLMs are utilized to generate synthetic text based on patterns learned from existing data. This method can produce diverse and realistic examples, but care must be taken to avoid perpetuating biases present in the original data.
    \item Hybrid Approaches: These combine multiple methods, such as using rule-based generation to create initial examples, followed by model-based generation to refine and expand the synthetic dataset.
\end{itemize}

\section{Experiment Setup}
To tackle the dataset challenge, we adopted the best practices of large language model training to create a synthetic dataset representing real-world scenarios of real estate transactions. This dataset is used to adapt general-purpose LLMs to be legal experts in real estate transactions. We collected public datasets on real estate transactions in different states and prepared templates as contexts to generate example contracts.

We obtained authentic real estate transaction datasets from public sources, covering real estate transactions in New York City\cite {nycpropertydata}. Those datasets are in tabular formats, which can be easily processed and understood by machine learning algorithms, and the columns contain key attributes of real estate transactions, like city, lot number, transaction time, and prices.

To generate the synthetic contract, we prompt the LLM with the contract templates from each state and provide these ground-truth attributes as contexts. We designed a variety of questions and answers that can be used to evaluate the accuracy of information retrieval and reasoning tasks.

For clauses not available in the real transaction data, we adopted a rule engine as described in Figure \ref{fig:synthetic} to generate random contract terms and use them as ground truth answers to evaluate the model's reasoning capabilities.

\begin{figure}
\centering
\begin{tikzpicture}[mindmap, grow cyclic, every node/.style=concept, concept color=orange!40,minimum size=2cm,
	level 1/.append style={level distance=4cm,sibling angle=45,minimum size=1.5cm}]
	level 2/.append style={level distance=1cm,sibling angle=45,minimum size=0.5cm}]
\node{Synthetic Contract}
	child { node [concept color=yellow!30] {Financing Contingency}}
	child { node [concept color=yellow!30] {Inspection Contingency}}
	child { node [concept color=yellow!30] {Closing Cost}}
	child { node [concept color=yellow!30] {Deposit Structure}}
;
\end{tikzpicture}
\caption{Method to generate synthetic contracts} \label{fig:synthetic}
\end{figure}

\section{Model Training and Experiments}
While the field of natural language processing has been dominated by massive language models with hundreds of billions of parameters, our work takes a different approach. We leverage a smaller large language model (LLM), with a scale of a few billion parameters, exemplified by models like LLaMA-8B\cite{2023llama} and Phi-3\cite{abdin2024phi3}. This deliberate choice is motivated by several crucial factors that align with our application's specific requirements and constraints. Firstly, deploying a smaller LLM significantly enhances privacy. Smaller models enable on-device processing, unlike their larger counterparts, which often necessitate cloud-based inference. This means sensitive user data remains on the user's device, minimizing the risks associated with data transmission and storage on external servers. This is particularly critical in privacy-sensitive domains where data security is paramount. Secondly, smaller LLMs offer practical advantages in inference speed and fine-tuning.  Their reduced scale translates to lower computational demands, making them easier to adapt and specialize using modest resources. This efficiency is particularly beneficial when working with synthetic datasets, as it allows for rapid experimentation and iterative refinement of the model to achieve optimal performance on the specific task.

To further enhance the efficiency and performance of our chosen LLM, we employ Parameter-Efficient Fine-Tuning (PEFT) techniques\cite{peft}. PEFT methods are designed to adapt large language models for specific tasks by fine-tuning only a small subset of parameters while keeping most of the model's weights frozen. By limiting the number of trainable parameters, PEFT significantly reduces the computational resources required for fine-tuning. This makes it feasible to adapt large models even with limited hardware. PEFT methods minimize the memory footprint during fine-tuning, enabling the training process on devices with less memory capacity. It also preserves the majority of general knowledge existing in the pre-trained model parameters. Low-Rank Adaptation (LoRA)\cite{hu2022lora} inserts adapters with trainable rank decomposition matrices into each layer of the transformer model. In our work, we carefully evaluate different PEFT methods to identify the most effective approach for fine-tuning our chosen LLM on the synthetic dataset. By combining the advantages of a smaller LLM with the efficiency of PEFT techniques, we balance performance, resource utilization, and privacy.

On top of Llama-3.1-8B Instruct model\cite{llama3}, we adopt the following LoRA hyper-parameter configs for fine-tuning, which results in a massive reduction of trainable parameters to only 10 million.

\begin{quote}
    \begin{itemize}
    \item alpha=64,
    \item dropout=0.05,
    \item rank = 4
    \item target modules = [
        "q proj", "k proj", "v proj", "o proj",
        "gate proj", "up proj", "down proj",]
\end{itemize}
\end{quote}

The model is fine-tuned for 3 epochs with our synthetic training dataset of over three thousand contracts with over 3 million tokens. We sampled 100 contracts from the held-out dataset as a validation set to monitor the loss. We can observe that the model converges within 3 epochs and starts over-fitting at the end of the third epoch, as shown in Figure \ref{fig:loss-curve}. The model is trained on one NVIDIA A100 GPU, and the total training time is around 8 hours.

\begin{figure}
    \centering
    \includegraphics[width=0.8\linewidth]{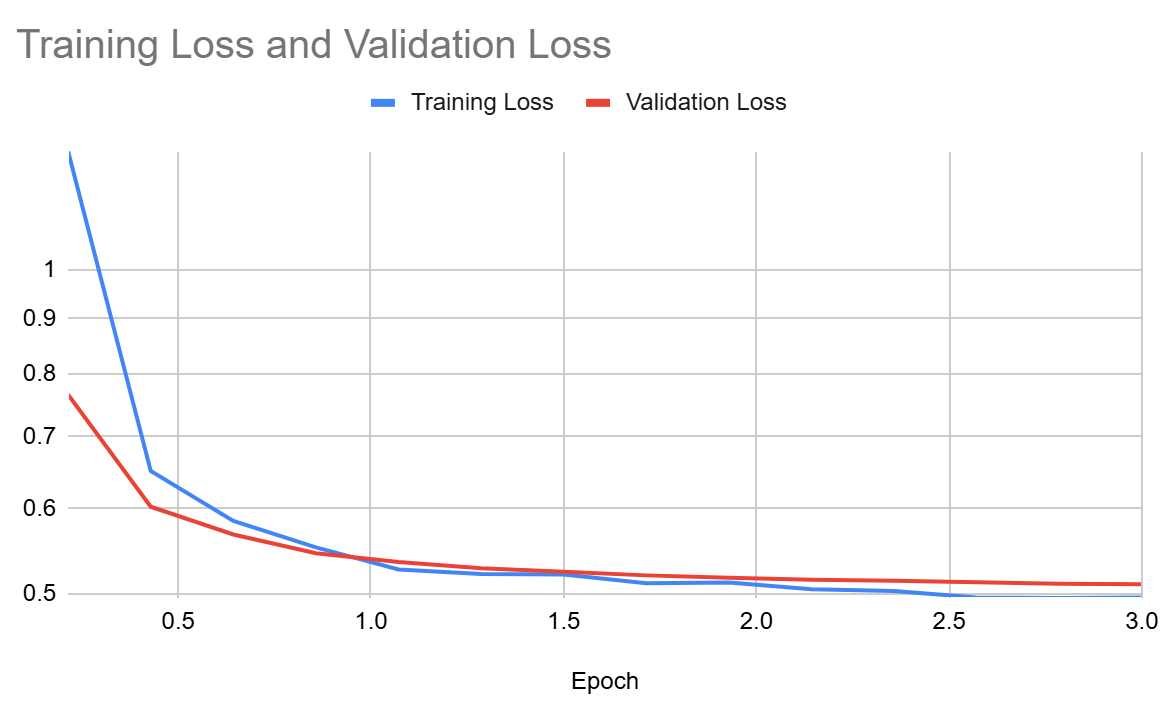}
    \caption{Fine-tuning can significantly reduce training loss and validation loss, and the validation loss converges at the end of epoch 3.}
    \label{fig:loss-curve}
\end{figure}

\section{Evaluation}
One of the most basic tasks that an LLM can perform is extracting information from a real estate contract. This includes information such as the buyer and seller's names, the property address, the purchase amount, and the closing date. This information can be used to create various reports and documents, such as title reports, closing statements, and tax returns.

In addition to extracting information from contracts, LLMs can also be used to perform more complex tasks, such as logical reasoning with scenario-based questions. For example, an LLM could be used to determine whether or not a seller can collect earnest money if the buyer cannot obtain financing. This type of task requires the LLM to understand the specific terms of the contract and apply them to a given scenario.

To reduce hallucination, we insert random questions about details unknown from the contract. This will help the LLM to learn that it cannot simply generate information that is not present in the contract. Here are some examples of random questions that could be used to reduce hallucination:
\begin{itemize}
    \item What is the property tax amount?
    \item What is the name of the insurance company providing the buyer's homeowners insurance?
    \item What is the lender's name providing the buyer's mortgage?
    \item What is the date of the buyer's loan application?
\end{itemize}

\subsection{Evaluation Metrics}
Evaluating the performance of our fine-tuned LLM on the question-answering (QA) task requires careful selection of appropriate metrics. While accuracy, measured as the percentage of correctly answered questions, provides a basic indication of performance, it fails to capture the nuances of language and the complexities of QA. Therefore, we employ a combination of metrics that provide a more comprehensive assessment.

Exact Match (EM) measures the proportion of answers that perfectly match the ground truth, emphasizing precise language understanding and generation. However, EM can be overly strict, especially for unstructured answers. To address this, we utilize the F1 score, which considers the overlap between predicted and ground truth answers at the token level. F1 score provides a more lenient evaluation, rewarding models that capture the essential information even with minor variations in phrasing.

Although the F1 scores provide a quantitative measure of answer accuracy, they often fall short of capturing the semantic similarity between the predicted and ground truth answers. To address this limitation, we incorporate BERTScore\cite{bert-score} as an evaluation metric for our QA system. Unlike string-based comparisons, BERTScore leverages pre-trained contextual embeddings from BERT to compute the similarity between two sentences. This allows for a more nuanced assessment that considers the semantic meaning and contextual information embedded within the text, rather than relying solely on lexical overlap. By utilizing BERTScore, we aim to evaluate the quality of generated answers based on their semantic alignment with the ground truth, even if they exhibit variations in phrasing or lexical choices. This approach provides a more comprehensive and meaningful evaluation of the LLM's ability to understand and respond to questions accurately and comprehensively.

Our evaluation demonstrates that, after fine-tuning, LLM is able to better match the reference answer in our synthetic dataset, with precision and recall improvements in BERT Score. The only regression in text-matching recall metrics is because the pre-trained model, without fine-tuning, tends to generate verbose answers and include sentences of the original contract, not providing the direct answer.

\begin{table*}[t]
\caption{Evaluation metrics show significant improvements after fine-tuning LLM with synthetic dataset}

\centering
\begin{tabular}{|ll|p{4cm}|p{4cm}|}
\hline
\multicolumn{2}{|l|}{Evaluation Metrics}                                                 & llama-3-8b-instruct \newline (Baseline) & After fine-tuning \\ \hline
\multicolumn{1}{|l|}{\multirow{4}{*}{BERT Score}}    & Precision                         & 0.638                            & 0.827             \\ \cline{2-4} 
\multicolumn{1}{|l|}{}                               & Recall                            & 0.833                            & 0.930             \\ \cline{2-4} 
\multicolumn{1}{|l|}{}                               & F1-Score                          & 0.722                            & 0.874             \\ \cline{2-4} 
\multicolumn{1}{|l|}{}                               & Match Rate ( F1 \textgreater 0.7) & 0.630                            & 0.968             \\ \hline
\multicolumn{1}{|l|}{\multirow{4}{*}{Text Matching}} & Precision                         & 0.099                            & 0.788             \\ \cline{2-4} 
\multicolumn{1}{|l|}{}                               & Recall                            & 0.914                            & 0.596             \\ \cline{2-4} 
\multicolumn{1}{|l|}{}                               & F1-Score                          & 0.168                            & 0.633             \\ \cline{2-4} 
\multicolumn{1}{|l|}{}                               & Match Rate (F1 \textgreater 0.5)  & 0.041                            & 0.613             \\ \hline
\end{tabular}
\end{table*}

\subsection{Qualitative Analysis}
We also provide some examples demonstrating the pros and cons of fine-tuning models.

\subsubsection{Good Examples}

Example 1: After fine-tuning, LLM is able to provide succinct and more direct and rephrased answers which is more understandable.
\begin{quote}
\textit{\textbf{Question:}} What happens if the Purchaser is not satisfied with the inspection?

\item \textit{\textbf{Answer (Baseline): }} According to Section 3(a) of the contract, if the Purchaser is not satisfied with the condition of the Premises after conducting inspections by the specified date, they have the right to terminate this Contract by providing written notice to the Seller prior to the expiration of the inspection period.
\item \textit{\textbf{Answer (Fine-tuned): }}  Upon termination, the escrow deposit of \$22,325 will be returned to the Purchaser, and all parties will be released from further obligations under this Contract.
\end{quote}

Example 2: The pre-trained model tends to generate irrelevant details to the answers.
\begin{quote}
\textit{\textbf{Question:}} What will the seller convey to the purchaser?

\item \textit{\textbf{Answer (Baseline): }} According to the contract, the seller (Ernestina Parker) shall sell and convey the property, together with all buildings and improvements thereon, to the purchaser (Jay Aschenbrenner). This includes the property described in Schedule A, which is 189 Schermernhorn Street, along with any rights to land lying in the bed of any street or highway adjoining the property.

\item \textit{\textbf{Answer (Fine-tuned): }} The property and all buildings and improvements thereon.
\end{quote}

Example 3: For terms unspecified in the contract, the baseline model generates answers with blank lines, without direct answers to the question.

\begin{quote}
\textit{\textbf{Question:}} Who holds the earnest money deposit in escrow?

\item \textit{\textbf{Answer (Baseline): }} According to paragraph 6, the earnest money deposit is held in escrow by \rule{1cm}{0.15mm} at \rule{1cm}{0.15mm}, in a segregated, interest-bearing account for the benefit of the parties.

\item \textit{\textbf{Answer (Fine-tuned): }} Escrow holder. (Name and address to be filled)
\end{quote}

\subsubsection{Failure Cases}
The fine-tuned model sometimes generates overly simplified answers, missing enough details.
\begin{quote}
\textit{\textbf{Question:}} Which costs is the Seller responsible for?

\item \textit{\textbf{Answer (Baseline): }}  Real estate agent commissions, transfer taxes, escrow fees
According to Section 5 of the contract, the Seller is responsible for the following costs:
Real estate agent commissions, Transfer taxes, All escrow fees associated with the Closing

\item \textit{\textbf{Answer (Fine-tuned): }} Seller's Costs.
\end{quote}

The fine-tuned model sometimes doesn't provide a direct answer.
\begin{quote}
\textit{\textbf{Question:}} Is the property being sold 'as is'?

\item \textit{\textbf{Answer (Baseline): }}  Yes, according to Clause 3, the Inspection Contingency, the property is being sold "as is" without any inspection contingency. This means that the Purchaser is assuming the property in its current condition, without the ability to conduct inspections or request repairs before closing.

\item \textit{\textbf{Answer (Fine-tuned): }} No inspection contingency. Purchaser assumes property "as is".
\end{quote}

\section{Future Directions}

The application of large language models in real estate contract analysis opens up several venues for future research and development:

\begin{itemize}
    \item \textbf{Multi-lingual Support:} Extending models to handle contracts in different languages to facilitate global real estate transactions.
    \item \textbf{Image Support:} Extending models to handle inspections and appraisal reports with images.
    \item \textbf{Pricing Guidance:} Extending the models to guide pricing. 
    \item \textbf{Regulatory Compliance:} Integrating legal compliance checks into automated contract analysis systems.
\end{itemize}

\section{Conclusion}

In this paper, we have discussed using large language models for extracting information from real estate sales contracts. By leveraging advanced NLP techniques, we can automate tedious tasks associated with contract analysis and improve efficiency in real estate transactions. Challenges such as legal complexity and ambiguity can be addressed using large language models and domain-specific fine-tuning. Future research focuses on enhancing multi-lingual support, semantic understanding, and regulatory compliance in automated contract analysis systems.

\bibliographystyle{ieeetr}
\bibliography{ms.bib}

\end{document}